\documentclass[sigconf,nonacm]{acmart} 

\AtBeginDocument{%
  }

\usepackage{graphicx}
\usepackage{xurl}
\usepackage{float}
\usepackage{array}
\usepackage{subcaption}
\usepackage{multirow}
\usepackage{tabularx}
\usepackage{amsmath}
\usepackage{dblfloatfix}

\renewcommand\footnotetextcopyrightpermission[1]{}

\acmConference[DAI 2026]
  {The 8th International Conference on Distributed Artificial Intelligence}
  {November 29--December 2, 2026}
  {Hong Kong}

\makeatletter
\renewcommand{\@authorfont}{\normalfont\large}
\renewcommand{\@affiliationfont}{\normalfont\fontsize{9.5}{11}\selectfont}
\makeatother

\begin{document}

\title[SMaRT-Tug]{SMaRT-Tug: Structured Multi-Agent Reinforcement Learning for Physics-Based Tugboat–Barge Collaborative Manipulation}
\settopmatter{authorsperrow=4}

\author{Junkai Lu}
\authornote{These authors contributed equally to this work.}
\affiliation{%
  \institution{National University of Singapore}
  \city{Singapore}
  \country{Singapore}
}
\email{jlu@nus.edu.sg}

\author{Jiadong Zhao}
\authornotemark[1]
\affiliation{%
  \institution{National University of Singapore}
  \city{Singapore}
  \country{Singapore}
}
\email{jzhao@nus.edu.sg}

\author{Jiacheng Zhang}
\affiliation{%
  \institution{National University of Singapore}
  \city{Singapore}
  \country{Singapore}
}
\email{e1554322@u.nus.edu}

\author{Wenqi Zhao}
\affiliation{%
  \institution{National University of Singapore}
  \city{Singapore}
  \country{Singapore}
}
\email{e1553489@u.nus.edu}

\author{Hao Gen Chia}
\affiliation{%
  \institution{ST Engineering Unmanned \& Integrated Systems Pte. Ltd.}
  \city{Singapore}
  \country{Singapore}
}
\email{haogen.chia@stengg.com}

\author{Qun Shen Png}
\affiliation{%
  \institution{ST Engineering Unmanned \& Integrated Systems Pte. Ltd.}
  \city{Singapore}
  \country{Singapore}
}
\email{qunshen.png@stengg.com}

\author{Germaine Ee}
\affiliation{%
  \institution{ST Engineering Unmanned \& Integrated Systems Pte. Ltd.}
  \city{Singapore}
  \country{Singapore}
}
\email{jiamingermaine.ee@stengg.com}

\author{Chengyang He}
\affiliation{%
  \institution{National University of Singapore}
  \city{Singapore}
  \country{Singapore}
}
\email{hechengyang@nus.edu.sg}

\author{Yifeng Zhang}
\affiliation{%
  \institution{National University of Singapore}
  \city{Singapore}
  \country{Singapore}
}
\email{yifengz@nus.edu.sg}

\author{Nathanael Tan}
\affiliation{%
  \institution{ST Engineering Unmanned \& Integrated Systems Pte. Ltd.}
  \city{Singapore}
  \country{Singapore}
}
\email{nathanael.tanez@stengg.com}

\author{Guillaume Sartoretti}
\affiliation{%
  \institution{National University of Singapore}
  \city{Singapore}
  \country{Singapore}
}
\email{guillaume.sartoretti@nus.edu.sg}

\renewcommand{\shortauthors}{J. Lu \& J. Zhao et al.}

\begin{abstract}
  Autonomous tugboating is central for automating maritime operations such as port logistics and vessel maneuvering, where multiple tugboats must cooperatively transport/manipulate a larger vessel. Collaborative pushing in this setting is challenging due to coupled hydrodynamics, low resistance, strong environmental disturbances, underactuated barge dynamics, and contact-rich interactions. Conventional control methods often rely on simplified models and fixed configurations, which limit their adaptability, while learning-based approaches are constrained by the lack of scalable and physically realistic training environments. We address these challenges by introducing a physics-based, GPU-accelerated simulation and learning framework for collaborative tugboat manipulation. Our simulator incorporates a customized buoyancy model, wave modeling, and hydrodynamic resistance, and supports large-scale multi-agent training under marine dynamics. In this simulator, we train a decentralized MAPPO (Multi-Agent PPO) policy augmented with a structured control prior (SCP) to improve training stability and maintain feasible pushing configurations. We evaluate our learned policy on straight-line transit, turning, and deceleration tasks, where we show that our decentralized framework yields more reliable and accurate maneuvering performance compared to a PID-based controller and a centralized PPO baseline. We further demonstrate zero-shot generalization to more challenging sea states and advanced maneuvers, as well as zero-shot scalability to larger teams of three and four tugboats despite training with only two agents.
\end{abstract}



\begin{teaserfigure}
  \centering
  \includegraphics[width=0.75\textwidth]{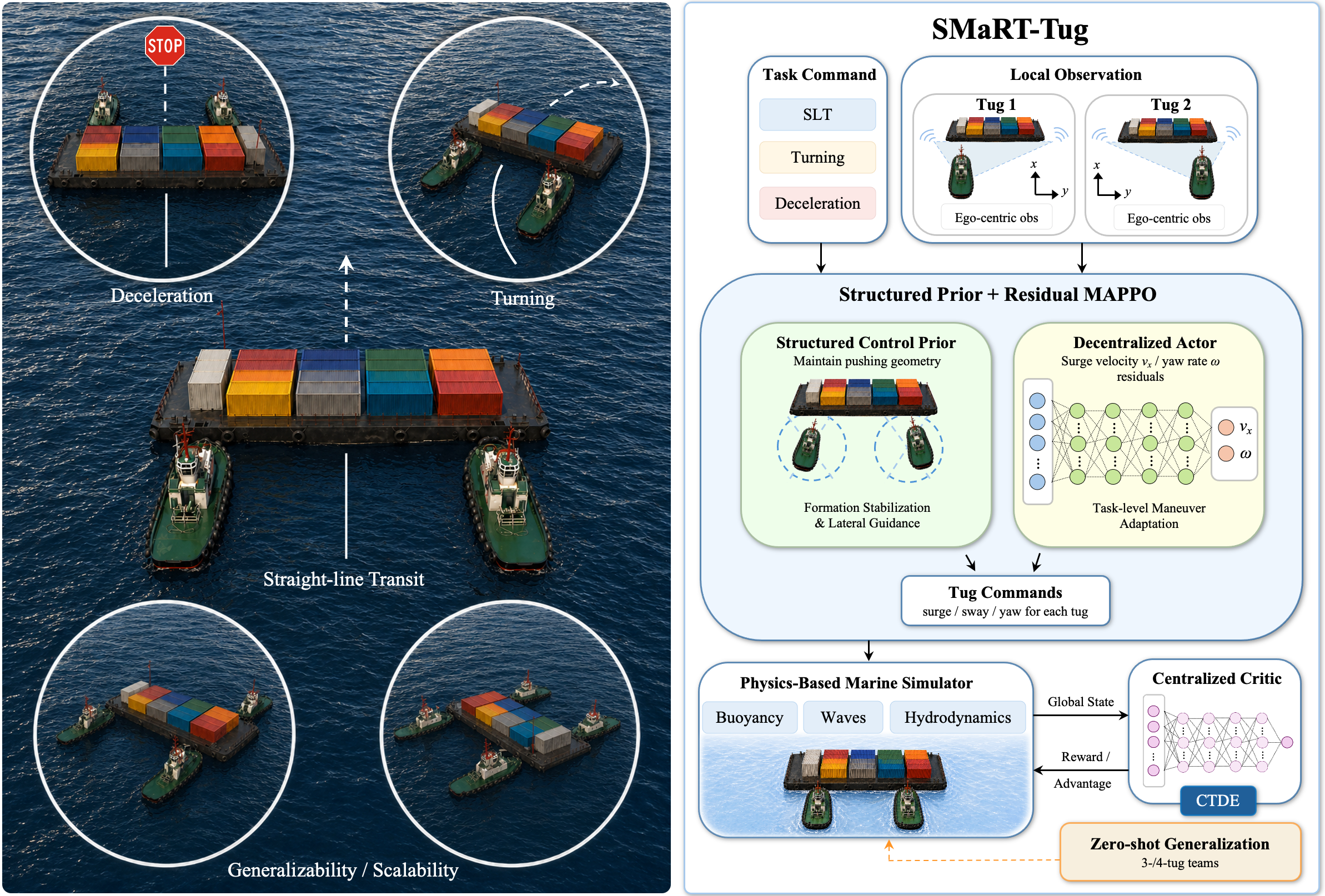}
  \caption{
    Overview of our proposed learning-based framework for decentralized
    collective tugboating, which relies on structured control priors (SCP)
    for general placement of the tugs and on a residual policy for object
    interaction, to achieve three key tasks: straight-line transit (SLT),
    turning, and deceleration. We further demonstrate the generalizability
    and scalability of our work across sea states and team sizes.
    \textsuperscript{*}
  }
  \Description{
    Overview of the tugboat--barge manipulation framework, including
    structured control priors, decentralized policy learning, maneuver
    tasks, and generalization across sea states and team sizes.
  }
  \label{fig:overview}
\end{teaserfigure}

\maketitle

\begingroup
\renewcommand{\thefootnote}{\fnsymbol{footnote}}
\footnotetext[1]{The left marine scene is based on snapshots from our simulator and was enhanced using AI-assisted recoloring and retexturing.}
\endgroup

\section{Introduction}

Autonomous tugboating plays an important role in maritime operations such as port logistics, vessel (un-)berthing, and offshore transportation~\cite{choi2023review}.
Among the different modes of tug-assisted vessel manipulation, contact-based pushing is particularly attractive in practice due to its direct interaction and relatively light deployment requirements.
Unlike single-vessel navigation, push-based tug assistance is a contact-rich multi-robot manipulation problem in which multiple tugboats must modulate their non-prehensile interactions with the barge to produce the desired barge motion.
Although cooperative transport and robotic pushing are both well-established topics in robotics, their overlap in multi-robot push-based manipulation remains relatively limited, especially for large floating bodies in marine environments~\cite{tuci2018cooperative,stuber2020let,tang2024collaborative}.
This motivates our study of collaborative pushing for autonomous barge maneuvering.

However, collaborative pushing is especially challenging in marine environments.
Surface vessels exhibit significant inertia, low resistance, and strong coupling among surge, sway, and yaw direction hydrodynamics, while environmental disturbances such as waves further complicate control.
These difficulties become even more pronounced in collaborative pushing, where the actions of multiple tugboats interact through both contact and hydrodynamics to determine the motion of the shared barge.
Existing studies address only parts of this problem.
Traditional marine control methods are effective for many single-vessel tasks and some tug-assisted operations, but they often rely on simplified models or fixed configurations that are difficult to generalize to contact-rich multi-tug manipulation of a shared underactuated barge, especially when team size or contact layout changes~\cite{fossen2000survey,karimi2021guidance,du2020cooperative,choi2020preliminary,barrera2021azimuth,lee2021optimization}.
Although learning-based methods have shown promise in related maritime tasks, work directly targeting decentralized multi-agent pushing of a shared vessel remains limited~\cite{ye2023deep,higo2023development,pereira2024reinforcement,wang2023deep,li2024design,xie2021reinforcement}.
Progress is further hindered by the lack of reinforcement-learning-oriented simulators that combine sufficient marine physics fidelity with the efficiency required for large-scale parallel training~\cite{lesy2025asvsim,newton_physics_engine,gan2025genesis,bingham2019toward,mittal2025isaac}.

To address these challenges, we first develop a physics-based simulation and learning framework for collaborative surface-vessel pushing, on top of NVIDIA's IsaacLab~\cite{mittal2025isaac}.
Our simulator incorporates voxelized buoyancy computation, wave modeling, and customized hydrodynamic resistance modeling, while still enabling efficient parallel simulation for reinforcement learning.
We then train a decentralized MAPPO (Multi-Agent PPO) policy~\cite{yu2022surprising} in this simulator.
There, to improve training stability and preserve feasible pushing behaviors, our learned policy is formulated as a residual controller over a structured control prior (SCP), implemented as a position-holding PID-based controller that maintains desirable relative position and orientation between the tugs and the barge.
With the SCP stabilizing the tugboats' positions, the policy focuses on the most complicated interaction dynamics rather than learning the entire behavior from scratch.
We further investigate zero-shot generalization by deploying the learned policy under more challenging sea states and in advanced maneuvers involving three and four tugboats, despite training only with two agents.
The resulting behaviors demonstrate robustness to stronger wave disturbances and scalability to larger teams, suggesting that the learned decentralized policy can generalize to both increased environmental disturbance and to expanded team size.
Together, these results highlight the promise of combining physically grounded simulation, structured control priors, and decentralized multi-agent learning for autonomous collaborative tugboat operations.

The main contributions of this paper are summarized as follows:

\begin{enumerate}
    \item We develop a physics-based, GPU-accelerated simulator for marine collaborative manipulation that enables parallel training, featuring realistic buoyancy via voxelization, wave modeling, and customized hydrodynamic resistance.
    \item We propose a multi-agent reinforcement learning framework that can effectively learn complex collaborative marine manipulation tasks by relying on structured control priors (SCP).
    \item We show that our learned policies can generalize to unseen sea states and scale to larger teams, highlighting their potential for real-world deployments in autonomous multi-agent marine tasks.
\end{enumerate}

\section{Related Work}
\subsection{Marine Vessel Control and Tug-assisted Maneuvering}
Marine vessel control has been extensively studied for maneuvering, path tracking, trajectory tracking, dynamic positioning, and berthing, using methods such as PID, nonlinear control, backstepping control, sliding mode control, robust control, and model predictive control (MPC)~\cite{fossen2000survey,karimi2021guidance}. 
These methods are interpretable, compatible with physical constraints and stability analysis, and are especially suitable for single-ship control. 
Existing research also includes tug-assisted maneuvering and autonomous tugboat operations for towing and berthing~\cite{choi2023review,du2020cooperative}. 
However, most studies still assume specific vessels, pre-defined tugboat configurations, or fixed operation scenarios, rather than treating multiple tugboats as autonomous intelligent agents that manipulate shared underactuated barges through physical interaction.

Many traditional methods rely on simplified models and explicit parametric modeling. The actual tug-barge interaction involves underactuation, disturbances, contact, friction, and time delays, but many works adopt point-contact assumptions, reduced dynamics, simplified tug-force models, or low-speed 3-DOF ship models for tractability~\cite{choi2020preliminary,barrera2021azimuth,lee2021optimization}. Model-based methods further require pre-given hydrodynamic coefficients, added mass, contact geometry, environmental forces, and thrust limits~\cite{barrera2021azimuth,lee2021optimization,bidikli2016robust}, so controllers often need redesign when vessel type, tug arrangement, contact topology, or disturbance models change. Scalability is also limited: many methods assume fixed tug numbers, role divisions, and operating ranges, while two-tug controllers and escort-tug force prediction methods are difficult to transfer to different team sizes, contact layouts, or angle ranges~\cite{lee2021optimization,du2020cooperative,aydin2018practical}. 
Therefore, although traditional maritime control provides an important foundation, contact-rich multi-tug barge handling still requires learning-based methods as a supplement to reduce reliance on precise analytical models and enhance adaptability across tasks, environments, and team sizes.

\subsection{Learning-based Control for Marine Cooperative Manipulation}
Learning-based control can complement model-driven methods in scenarios involving rich contact interactions, complex fluid dynamics, and multi-body coordination.
Existing studies show that deep learning and reinforcement learning have been applied to ship control, navigation, monitoring, and maritime logistics~\cite{ye2023deep}.

Learning has particularly great potential for low-speed maritime maneuvering with abundant contacts. 
Although learning-based research specifically targeting tugboat-assisted barge operation is still scarce, DDQN-based automatic berthing/unberthing has been validated on a large ferry under modeling errors and disturbances~\cite{higo2023development}, and RL-based autonomous surface vehicle (ASV) docking has been tested in simulated and real scenarios~\cite{pereira2024reinforcement}.
Meanwhile, floating-object manipulation can be classified as attaching, caging, pushing, and towing, indicating that push-based tugboat manipulation is a clear but still underexplored task category~\cite{du2023review}.

Learning can also reduce the reliance on precise analytical models under complex hydrodynamic conditions. 
The ASV tracking control based on deep reinforcement learning (DRL) has demonstrated competitive performance in natural water bodies and considers wind, waves, currents, measurement noise, and non-ideal actuators~\cite{wang2023deep}.
Actor-critic model-following control has also been implemented on azimuth stern drive (ASD) tugboats with model-scale scaling and physically deployed through simulation pre-training~\cite{li2024design}. These studies illustrate that learning is not a replacement for physical models but rather a means of learning complex relationships that are difficult to model analytically.

However, learning-based maritime operation directly related to the contact-rich manipulation of shared objects remains limited. 
Existing multi-agent maritime RL mainly focuses on non-contact collaborative scenarios, such as the generation and maintenance of multiple unmanned surface vehicle (USV) formations~\cite{xie2021reinforcement}. 
Imitation learning is also not a direct alternative: based on automatic identification system (AIS) data, imitation learning can generate berthing trajectories similar to those of a captain, but such data is specific to a particular task, at the trajectory level, and for a single vessel~\cite{higaki2025docking}. 
It cannot cover the multi-agent contact interaction distribution required for multi-tug coordinated operation.
Moreover, the lack of open-source, high-fidelity simulators and datasets remains a bottleneck in ~\cite{lesy2025asvsim}. 
Therefore, this paper proposes a physical simulation framework for multi-tugboat and barge operations and a structured multi-agent reinforcement learning method.

\subsection{Simulators for Multi-robot Learning in Physics-rich Environments}

Differentiable simulators such as Newton~\cite{newton_physics_engine} and Genesis~\cite{Genesis} provide significant extensibility and broad physical modeling capabilities. Nevertheless, they are positioned as general-purpose simulators rather than dedicated platforms for collaborative marine manipulation. 
In particular, adapting their particle-based fluid simulation, such as Smoothed Particle Hydrodynamics (SPH) or Material Point Method (MPM), into training scenarios that involve buoyancy, hydrodynamic drag, wave interaction, and multi-vessel coordination may introduce considerable computational overhead in large-scale RL or MARL training environments.

A second category comprises general-purpose simulation and training engines such as Unity~\cite{juliani2018unity}, Unreal~\cite{epic_learning_agents}, and UNIGINE~\cite{unigine_sim_sdk}, which provide mature ecosystems for scene construction, visualization, and learning workflows. However, these platforms are typically not designed as GPU-native batched robotics RL simulators, in contrast to IsaacLab, which integrates both physics simulation and policy training entirely on GPU~\cite{mittal2025isaac}.

A third category includes maritime-specific simulators such as K-Sim and VRX, which provide stronger domain realism for surface-vessel research through hydrodynamics, buoyancy, waves, and related vessel-level modelling~\cite{ksim_navigation,bingham2019toward}.
Open-source marine toolkits such as MultiVessel\_Simulation and asv\_wave\_sim further extend this space toward multi-vessel scenarios and wave-aware vessel dynamics~\cite{multivessel_simulation,asv_wave_sim}. 
However, these platforms are primarily focused on relatively constrained benchmark evaluation or more limited simulation environments, rather than unified environments for open-ended, massively parallel RL or MARL in cooperative tugboat-barge transport and manipulation~\cite{ksim_navigation,bingham2019toward,multivessel_simulation,asv_wave_sim}.

\section{Problem Formulation}
\label{sec:problem-formulation}
\vspace{0.2cm} \noindent \textbf{System Setup. }
We consider collaborative tugboat-barge maneuvering as a decentralized multi-agent cooperative manipulation problem, in which multiple (here, two at first) tugboats physically interact with a single barge to realize commanded planar maneuvers. 
The barge is the manipulated object, while the tugboats act as the controlled agents.
Unlike standard single-vessel navigation, this setting combines persistent contact interaction, coupled marine dynamics, and coordination through a common underactuated object.
In particular, the barge exhibits substantial inertia, limited controllability, and strong coupling between translational motion and yaw, rendering low-speed maneuvering nontrivial.

In general, the team is tasked with continuously tracking a (potentially time-dependent) commanded barge motion in the horizontal plane.
Under this formulation, straight-line transit (SLT), turning, and deceleration are treated as different task instances of a common cooperative manipulation problem, rather than as unrelated control tasks.

\vspace{0.2cm} \noindent \textbf{Observation, Action, and Reward. }
We formulate the cooperative tugboat-barge manipulation task as a decentralized partially observable Markov decision process (Dec-POMDP), where each tugboat makes decisions based on ego-centric local observations while jointly influencing the global barge dynamics through contact-rich physical interactions.
Each tugboat is modeled as a decentralized agent indexed by \(i \in \{1,2\}\). At decision step \(t\), each agent receives a local observation \(o_i^t\) and outputs an action \(a_i^t\) according to
\begin{equation}
a_i^t = \pi_{\theta_i}(o_i^t), \qquad i \in \{1,2\},
\end{equation}
where \(\pi_{\theta_i}\)  denotes the policy of agent \(i\). 
In our implementation, we adopt a parameter-sharing framework in which all tugboat agents use common policy parameters, i.e., \(\theta_i = \theta\), to improve sample efficiency and scalability. 
An agent role indicator is included in the observation to preserve agent-specific information. The actor observation is strictly local and ego-centric, including the observing tugboat's last command, the barge velocity, the tug-barge relative pose and velocity, the current task command, an agent role indicator, and a short history of recent features. During training, the critic uses a self-first global state that additionally includes both tugboats and privileged physics parameters. This yields a centralized-training-decentralized-execution (CTDE) formulation. 
The action is a compact tug-local command consisting of a forward velocity component and a yaw-rate component, while lateral correction is provided by the SCP. The two agents are trained with a shared return where the reward encourages command tracking, stable tug-barge geometry, and
safe termination behavior. The full observation, action, and reward specifications are provided in Appendix~\ref{app:policy_spec}.

\section{Physics-Based Simulation Framework}
\label{sec:simulation-framework}
This section presents our physics-based simulator developed to support physically grounded and computationally efficient learning of cooperative tugboat manipulation. Marine interaction is inherently challenging due to coupled hydrodynamics, low resistance, and complex wave disturbances, which are often simplified or neglected in existing reinforcement learning environments.

Our simulator is designed to balance physical fidelity and training scalability, capturing the key dynamics that govern surface vessel interaction while enabling large-scale parallel training. The framework integrates a GPU-accelerated simulation engine with three main modeling components: a voxel-based buoyancy model together with a realistic wave model for hydrostatic force computation, as well as a calibrated hydrodynamic resistance model that captures the anisotropic drag effects.

\vspace{0.2cm} \noindent \textbf{Simulation Engine. }
Our simulator is built on top of IsaacLab~\cite{mittal2025isaac}, a GPU-accelerated rigid-body physics engine for efficient large-scale simulation. All vessels are modeled as rigid bodies with contact interactions resolved by the underlying physics solver, while we incorporated custom forces, including buoyancy and water resistance to capture marine-specific dynamics. 

Our simulator is designed for parallel reinforcement learning, with all calculations executed in batch on the GPU. It vectorizes heavy wave surface queries, enabling simultaneous buoyancy calculations across environments.

\vspace{0.2cm} \noindent \textbf{Wave Model. }
To model environmental disturbances efficiently, we use a simplified version of Gerstner-wave height field~\cite{von1804theorie,fournier1986simple}:
\begin{equation}
\eta(\mathbf{p}, t)
=
z_w
+
\frac{A}{N_d}
\sum_{j=1}^{N_d}
\alpha_j
\sum_{m=1}^{N_\omega}
a_m
\cos\!\left(
k_m \mathbf{d}_j^T \mathbf{p}
-
\omega_m t
+
\phi_m
\right),
\end{equation}
where $\eta(\mathbf{p},t)$ is the water height at horizontal position $\mathbf{p}=(x,y)^T$, $z_w$ is the mean water level, and the remaining terms define the amplitude, direction, frequency, and phase of the wave components. By neglecting horizontal particle displacement, the water surface becomes an explicit height field that can be directly queried for batched buoyancy computation in parallel training. Further details of the wave generation model are provided in Appendix~\ref{app:wave}.

\vspace{0.2cm} \noindent \textbf{Buoyancy Calculation. }
Buoyancy is computed using a voxel-based approximation of each vessel's geometry. The vessel body is discretized into a set of volumetric cuboids, and the submerged voxels are identified by comparing their center positions with the local wave height. The displaced volume is then approximated from the total volume of submerged voxels, and the resulting buoyancy is applied at the corresponding center of buoyancy.

This formulation provides an efficient approximation of Archimedes' principle while remaining well suited for GPU-based batched simulation. It also allows the voxel resolution to be adjusted to trade off computational cost and accuracy. The detailed buoyancy equations are provided in Appendix~\ref{app:buoyancy}.

\vspace{0.2cm} \noindent \textbf{Hydrodynamic Resistance Model. }
To capture the low-speed planar response of the manipulated barge during reinforcement learning, we equip the barge with an explicit hydrodynamic resistance model in the RL-oriented physics-based simulator. This model provides a physically grounded and computationally efficient representation of low-speed barge maneuvering. In the present training setup, the model is applied only to the barge. The tugboats are velocity-controlled through simulator-side filtering of commanded velocities and therefore do not use the same explicit hydrodynamic resistance model.

For the barge, the hull hydrodynamics follow the unified hull-force formulation~\cite{yoshimura2009unified}, in which the surge, sway, and yaw components are represented by a combination of linear hydrodynamic derivatives and cross-flow drag terms. The corresponding non-dimensional hull-force expressions are
\begin{equation}
X_H' = - X_0' u' + \left( m_y' + X_{vr}' \right) v' r',
\end{equation}
\begin{equation}
Y_H' = Y_v' v' \left| u' \right| + \left( Y_r' - m_x' \right) r' u'
- C_D \int_{-0.5}^{0.5} \left( v' + C_{rY} r' x \right)
\left| v' + C_{rY} r' x \right| \, \mathrm{d}x,
\end{equation}
\begin{equation}
N_H' = N_v' v' u' + N_r' r' \left| u' \right|
- C_D \int_{-0.5}^{0.5} x \left( v' + C_{rN} r' x \right)
\left| v' + C_{rN} r' x \right| \, \mathrm{d}x.
\end{equation}
The detailed hydrodynamic resistance formulation, including our real-world water-tunnel-based resistance calibration, is provided in Appendix~\ref{app:Hydrodynamics Resistance Model}.

\section{Structured Multi-Agent Learning Framework}
\vspace{0.2cm} \noindent \textbf{Structured Control Prior. }
Direct end-to-end learning is difficult because the policy must simultaneously discover feasible cooperative pushing geometry and task-directed barge maneuvering under coupled marine dynamics. To reduce this burden, we introduce a structured control prior (SCP), a PID controller in the barge reference frame. For each tugboat, the SCP regulates a desired relative position and alignment with respect to the barge and produces a nominal tug-local velocity reference. Its role is not to solve the full maneuver task, but to keep the tugboats near designated contact regions and physically meaningful cooperative configurations from which task-level adaptation becomes tractable.

In the present implementation, the SCP is realized as a velocity-space proportional prior based on relative pose and alignment feedback. It provides nominal geometric regulation and, in particular, the lateral correction required to maintain cooperative contact and alignment during pushing, turning, and stopping. 

\vspace{0.2cm} \noindent \textbf{Residual Policy Learning. }
On top of the SCP, we learn a residual multi-agent policy. Each tugboat receives a strict ego-centric local observation and outputs a compact tug-local command. Instead of controlling the full local velocity vector, the policy acts primarily on forward motion and yaw regulation, while the lateral channel is supplied by the SCP. Let $o_i$ denote the local observation of tugboat $i$, $\Delta \mathbf{a}_i=\pi_\theta(o_i)$ the learned residual action, and $\bar{\mathbf{u}}_i$ the nominal base command. The executed control is
\begin{equation}
\mathbf{u}_i = \mathcal{G}\!\left(\bar{\mathbf{u}}_i, \Delta \mathbf{a}_i\right),
\end{equation}
where $\mathcal{G}(\cdot)$ combines the nominal command with the learned residual correction. In our implementation, the SCP supplies the lateral correction and part of the nominal geometric regulation, while the learned policy primarily modulates forward speed and yaw rate.

The learning framework follows standard MAPPO with parameter sharing across the two tugboats. Each actor receives only local observations together with an agent role indicator, whereas the critic uses a privileged centralized state representation during training (and is unused during deployment). To preserve agent-specific structure, centralized critic inputs are arranged in a self-first manner, so each tug evaluates the shared team behavior from its own perspective while still exploiting global coordination information.

\vspace{0.2cm} \noindent \textbf{Training Protocol. }
Training is conducted in batched parallel simulation using shared-return MAPPO under the CTDE setting described above. All policies are
trained with 128 parallel simulation environments. The primary training setup targets the two-tugboat manipulation problem, where both agents are optimized jointly toward the same barge-level maneuvering objective. To expose the policy to a broad range of contact and hydrodynamic interaction patterns, the training distribution spans commanded maneuvers, initial conditions, and selected task-level and physical variations. When stronger specialization or improved convergence is needed, the same framework also supports continued training from existing checkpoints and moderate task-difficulty curricula. These staged procedures improve optimization stability and sample efficiency in the contact-rich, dynamically coupled marine setting considered here, while leaving the underlying learning formulation unchanged.

\section{Experiments}
\label{sec:experiments}

\subsection{Experimental Setup}

\vspace{0.2cm} \noindent \textbf{Basic Tasks. }
We consider three groups of basic tasks that capture the core capabilities required for cooperative barge maneuvering.

\paragraph{Straight-Line Transit (SLT)}
The barge is required to track a commanded forward velocity (forward refers to the right side or starboard side of the barge) while maintaining the desired orientation. \textbf{\hypertarget{text-task-a}{\hyperlink{tab-task-a}{Task A}}} is to verify the deviation between the actual forward velocity and the command forward velocity by calculating the mean square error (MSE, lower is better). \textbf{\hypertarget{text-task-b}{\hyperlink{tab-task-b}{Task B}}} observes recovering ability from initial velocity-direction deviations, based on the maximum recoverable velocity deviation angle (higher is better).

\paragraph{Turning. }
The barge is required to rotate to a target yaw angle such that its final heading and velocity direction align with the command. \textbf{\hypertarget{text-task-c}{\hyperlink{tab-task-c}{Task C}}} assesses the turning maneuverability of the barge when it makes a $90^\circ$ turn, based on the lateral displacement when turning finishes (lower is better). \textbf{\hypertarget{text-task-d}{\hyperlink{tab-task-d}{Task D}}} assesses the ability to recover from lateral velocity induced by hydrodynamics after a $30^\circ$ turn, based on whether the recovery is successful and the deviation during the recovery process (lower is better).

\paragraph{Deceleration. }
Starting from a reverse-moving state, the barge is required to decelerate to rest. \textbf{\hypertarget{text-task-e}{\hyperlink{tab-task-e}{Task E}}} evaluates the ability to suppress the remaining speed in the moving direction during the deceleration process based on the absolute value of the residual longitudinal ($y$-axis) speed after deceleration (lower is better). \textbf{\hypertarget{text-task-f}{\hyperlink{tab-task-f}{Task F}}} assesses the adaptability in the presence of an angled initial speed, measuring the absolute value of the residual lateral ($x$-axis) velocity after deceleration (lower is better).

\vspace{0.2cm} \noindent \textbf{Full Maneuver.} 
The barge is required to complete a waypoint-following mission that combines straight-line transit (SLT), turning, deceleration, and reposition within a single episode, evaluating whether the policy can sustain stable control and perform smooth transitions across consecutive maneuvering phases. In particular, repositioning is handled by predefined trajectory tracking rather than a learned policy.

\vspace{0.2cm} \noindent \textbf{Baselines. }
We compare our MAPPO-based policy against a PID-based controller (proportional-only in our case) and a centralized PPO baseline (one policy controlling both tugboats). For the PID-based controller, since the role of the yaw-rate action during contact-based pushing is difficult to model explicitly, we only control the tugboats' longitudinal velocity commands. For each basic task, we define a task-specific set of proportional gains, consisting of three terms that regulate the barge velocity magnitude, velocity direction, and orientation error, respectively. For the detailed gain settings, see Table~\ref{tab:pid_gains} in Appendix~\ref{sec:exp_setting}.

\begin{figure*}[!b]
    \centering
    \includegraphics[width=\linewidth]{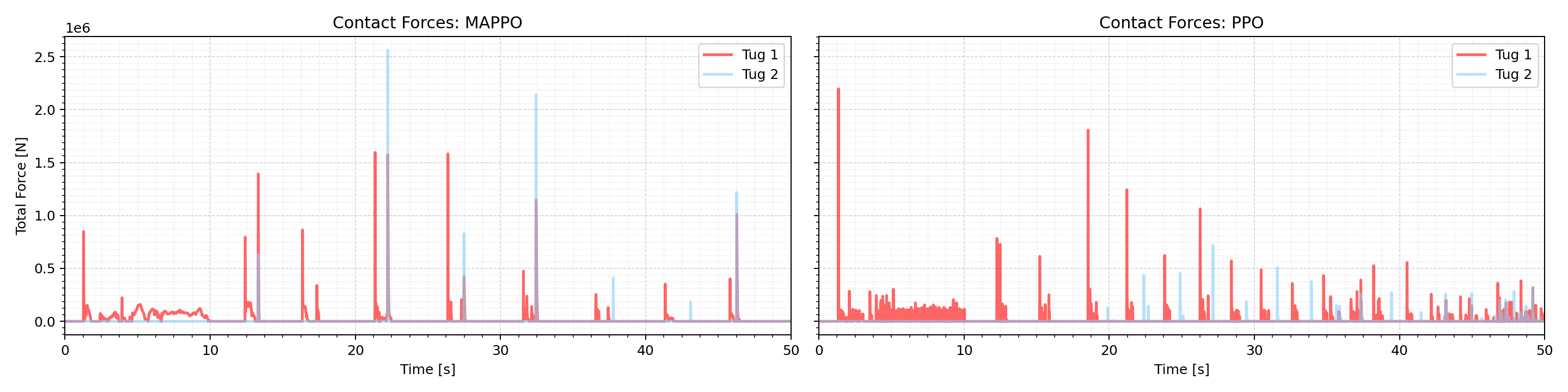}
    \caption{Contact force patterns comparison between PPO and MAPPO throughout \textbf{Task D}.}
    \label{fig:turning_force}
\end{figure*}

\subsection{Results}
\vspace{0.2cm} \noindent \textbf{Basic Tasks. }
\begin{table*}[t]
\centering
\caption{Performance comparison on basic maneuver tasks.}
\label{tab:basic-task}
\begin{tabularx}{\linewidth}{c|l|>{\centering\arraybackslash}X>{\centering\arraybackslash}X>{\centering\arraybackslash}X}
\hline
Category & \multicolumn{1}{c|}{Task} & PID & PPO & MAPPO \\ \hline

\multirow{2}{*}{SLT}
& \hyperlink{text-task-a}{A} - Velocity MSE $\downarrow$ ($\times 10^{-4}\,m^2/s^2$) 
& 213 & \textbf{4.39} & \underline{9.05} \\

& \hyperlink{text-task-b}{B} - Max Recovery Angle $\uparrow$ ($^\circ$)
& 5 & \underline{20} & \textbf{80} \\ \hline

\multirow{3}{*}{Turning}
& \hyperlink{text-task-c}{C} - $90^\circ$ Lateral Displacement $\downarrow$ ($m$)
& 122.29 & \underline{111.27} & \textbf{111.09} \\

& \hyperlink{text-task-d}{D} - Recovered from $30^\circ$ Turn
& No & \textbf{Yes} & \textbf{Yes} \\

& \hyperlink{text-task-d}{D} - $30^\circ$Lateral Displacement $\downarrow$ ($m$)
& N/A & \underline{115.1} & \textbf{37.3} \\ \hline

\multirow{2}{*}{Deceleration}
& \hyperlink{text-task-e}{E} - $|$Y Residual Velocity$|$ $\downarrow$ ($\times 10^{-2}\,m/s$)
& \underline{8.58} & 37.1 & \textbf{3.14} \\

& \hyperlink{text-task-f}{F} - $|$X Residual Velocity$|$ $\downarrow$ ($\times 10^{-1}\,m/s$)
& 2.49 & \underline{2.32} & \textbf{1.81} \\ \hline

\end{tabularx}
\end{table*}
The results in Table~\ref{tab:basic-task} highlight performance differences across the three groups of tasks. In the \textbf{Straight-Line Transit} (SLT) tasks, learning-based methods significantly outperform the PID-based controller in velocity tracking, with MAPPO achieving strong robustness to initial misalignment, supporting up to $80^\circ$ deviation compared to only $5^\circ$ for the baseline. In the \textbf{Turning} tasks, for large-angle maneuvers, all methods adopt an overshoot-and-correct behavior to complete the turn and overcome the lateral velocity; this task evaluates pure turning performance. The lateral displacement turns out comparable, although MAPPO still achieves the lowest error. In smaller-angle turning, where there is no overshoot-and-correct, this emphasizes the recovery ability after turning. MAPPO is able to directly suppress the lateral velocity induced during the maneuver, leading to substantially reduced displacement compared to PPO. To further analyze this behavior, we compare the contact forces generated by MAPPO and PPO in Fig.~\ref{fig:turning_force}. After the first \(10\,\mathrm{s}\) of the pure turning phase, both controllers start to suppress the lateral velocity and recover toward the commanded motion. The learned coordination patterns, however, differ substantially. MAPPO exhibits more synchronized contact, with force peaks from the two tugboats often occurring at the same time. By contrast, PPO produces a more alternating pattern, in which significant contact force is typically applied by only one tugboat at each timestep. Combined with the quantitative results, this indicates that synchronized contact plays an important role in fast recovery, and suggests that MAPPO learns a more cooperative strategy that improves recovery performance. In the \textbf{Deceleration} tasks, MAPPO consistently achieves the lowest residual velocities under both evaluation settings, indicating effective suppression of coupled motion components. The conventional controller partially reduces lateral velocity but struggles with cross-axis residuals, while PPO exhibits larger remaining velocities. Overall, these results highlight that our MAPPO policy yields superior accuracy and adaptability across all tasks.

\begin{figure*}[t]
    \centering
    \includegraphics[width=\linewidth]{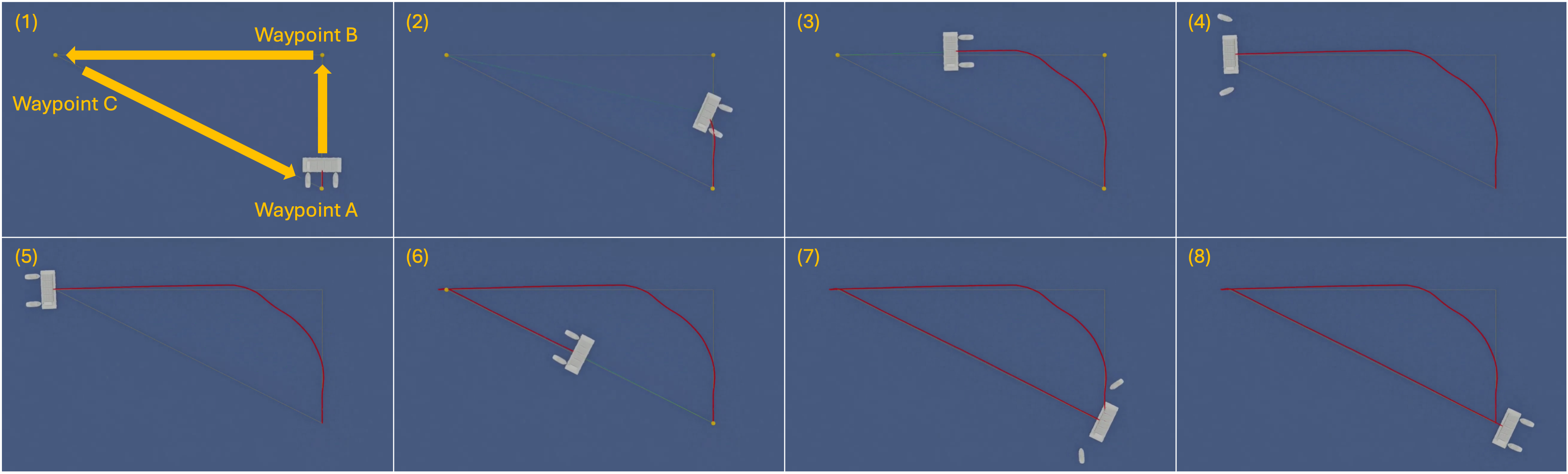}
    \caption{In the Full Maneuver task, the barge is commanded to follow a three-waypoint trajectory, from A to B to C. The trajectory is executed following eight stages: (1)~SLT, (2)~turn left, (3)~SLT, (4)~reposition, (5)~decelerate, (6)~turn right and SLT, (7)~reposition, and (8)~decelerate.}
    \label{fig:full_maneuver}
\end{figure*}

\vspace{0.2cm} \noindent \textbf{Full Maneuver. }
In the \textbf{Full Maneuver} task, the system successfully executes the multi-stage sequence using a simple high-level controller that switches between stage-specific policies (Fig.~\ref{fig:full_maneuver}), including SLT, turning, deceleration, and repositioning. Despite using a simple high-level controller without any smoothing mechanism, transitions between different stages remain stable and continuous. Each policy produces predictable end states that naturally fall within the initial state distribution of the subsequent stage, reducing the likelihood of out-of-distribution transitions during execution. Notably, even with a slight geometrical asymmetry in the barge, the execution of turning and SLT in stage 6 remains stable. The policy is able to perform pushing from the unseen side of the barge without having seen this scenario during training. Overall, the system reliably integrates basic skills into a coherent and robust full maneuver. This full maneuver capability is particularly meaningful for practical barge transit, as it enables navigation through complex port or marine environments using purely push-based interaction.

\vspace{0.2cm} \noindent \textbf{Ablation of SCP Strength. }
We compare multiple \(x\)- and yaw-axis gain settings of the SCP, while keeping the \(y\)-axis gain active because the learned policy does not output a \(y\)-axis action. Therefore, the following ablation ``SCP'' refers specifically to the \(x\)- and yaw-axis components. We also evaluate a decaying variant in which the gains are gradually reduced during training and set to zero at inference, following the idea of decaying action priors~\cite{sood2024decap}. Our decaying gain setting is designed to test whether the policy can follow the guidance provided by the SCP during early training, to learn how to maintain its relative position, and whether this capability can be learned when the SCP support is gradually removed. Detailed SCP gain settings are provided in Table~\ref{tab:controller_ablation} under Appendix~\ref{sec:exp_setting}. This ablation study examines how the strength of the SCP affects training stability and model performance.

\begin{figure}[t]
    \centering
    \includegraphics[width=0.9\linewidth]{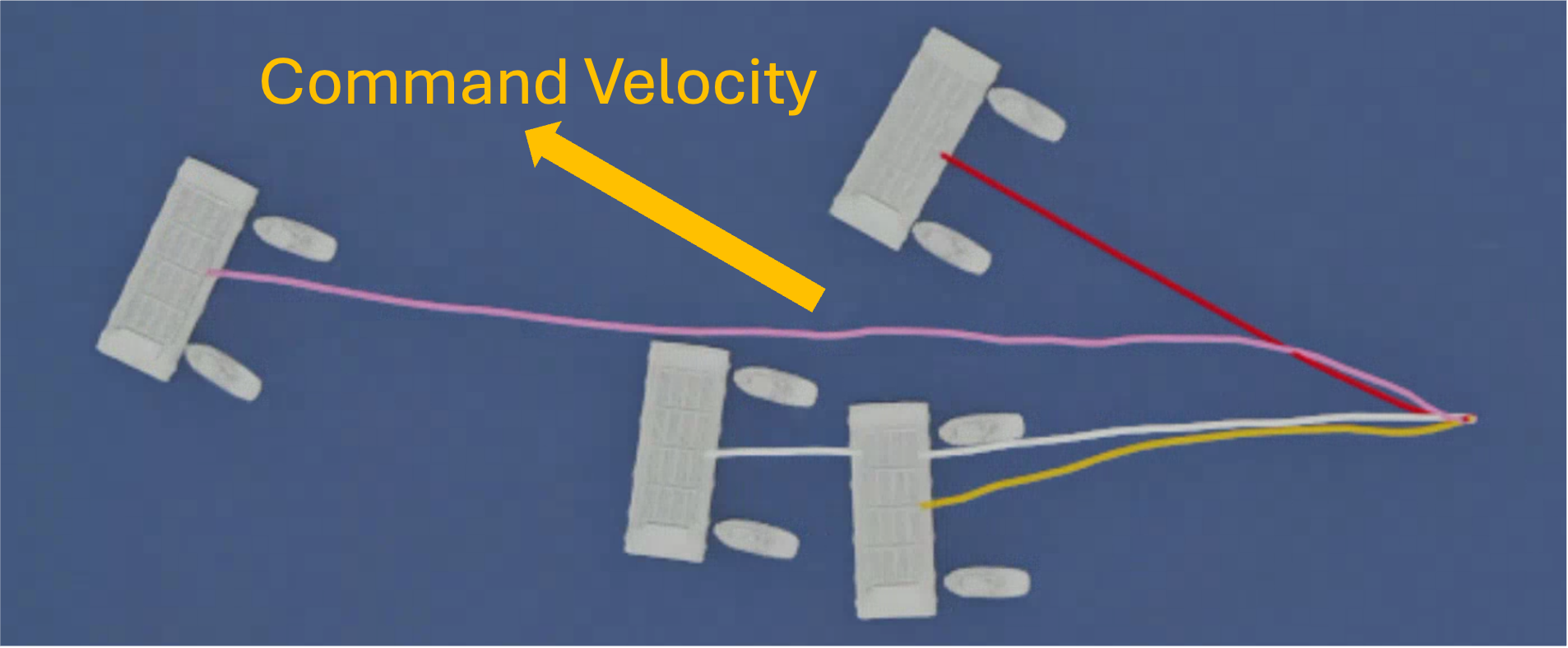}
    \caption{Ablation of SCP strength with $30^\circ$ command angle. Trajectories show the behavior of policies trained with full (red), weak (pink), no (white), and decaying (yellow) SCP.}
    \label{fig:ablation_pcontroller}
\end{figure}

The results of our SCP ablation are illustrated in Fig.~\ref{fig:ablation_pcontroller}. With the nominal SCP, the policy achieves stable and accurate control, quickly aligning both heading and velocity. When the SCP strength is reduced to half, the policy is still able to correct orientation, but fails to suppress the induced lateral velocity, resulting in persistent drift. In contrast, both zero SCP and decaying-SCP variants fail to converge to a usable policy. These results indicate that a sufficiently strong SCP is critical for stabilizing training, hence producing more effective policies.

\subsection{Generalizability and Scalability}
\vspace{0.2cm} \noindent \textbf{Effect of Training Wave Amplitude on Generalization. }
We train one policy with wave amplitude $0\,\mathrm{m}$ and another with wave amplitude $0.8\,\mathrm{m}$, and compare both policies under wave amplitudes $\{0, 0.4, 0.8, 1.2\}\,\mathrm{m}$ without retraining. Robustness is evaluated using the velocity-tracking MSE in the straight-line transit task (lower is better), averaged over 100 environments with randomized initial locations for each policy and wave amplitude.
The effect of wave amplitude during training on policy performance is summarized in Table~\ref{tab:ablation-wave-amp}. As wave amplitude increases, the free drift error rises significantly, indicating more challenging dynamics. The policy trained under calm conditions (0.0m) performs poorly even in the nominal setting, suggesting that the lack of environmental disturbance leads to suboptimal solutions, likely due to premature convergence to a local optimum. In contrast, the policy trained under moderate waves (0.8m) performs well across 0.0–0.8m conditions and remains more robust at 1.2m, despite some degradation. The 1.2m condition represents an extreme scenario that is unlikely in typical environments, under which all policies struggle; the 0.8m-trained policy still performs slightly better than the 0.0m-trained one, though. These results indicate that training with sufficient disturbance improves robustness and performance across varying environments.

\begin{table*}[t]
\centering
\caption{Effect of wave amplitude during training on policy robustness. Performance is evaluated under varying wave amplitudes using velocity tracking MSE.}
\label{tab:ablation-wave-amp}
\begin{tabularx}{\linewidth}{>{\centering\arraybackslash}p{5.5cm}|*{4}{>{\centering\arraybackslash}X}}
\hline
Wave Amplitude ($m$) & 0.0 $\downarrow$ & 0.4 $\downarrow$ & 0.8 $\downarrow$ & 1.2 $\downarrow$ \\ \hline

Free Drift MSE ($\times 10^{-4}\, m^2/s^2$) 
& 0.0 & 5.0 & 135.6 & 827.6 \\ \hline

\begin{tabular}[c]{@{}c@{}}
Trained at 0.0m MSE
($\times 10^{-4}\, m^2/s^2$)
\end{tabular}
& 756.9 & 2178.1 & 1895.8 & 2006.2 \\

\begin{tabular}[c]{@{}c@{}}
Trained at 0.8m MSE
($\times 10^{-4}\, m^2/s^2$)
\end{tabular}
& \textbf{8.5} & \textbf{13.2} & \textbf{221.4} & \textbf{1622.4} \\ \hline

\end{tabularx}
\end{table*}

\begin{figure}[!htbp]
    \centering

    \begin{subfigure}{0.7\linewidth}
        \centering
        \includegraphics[width=0.68\linewidth]{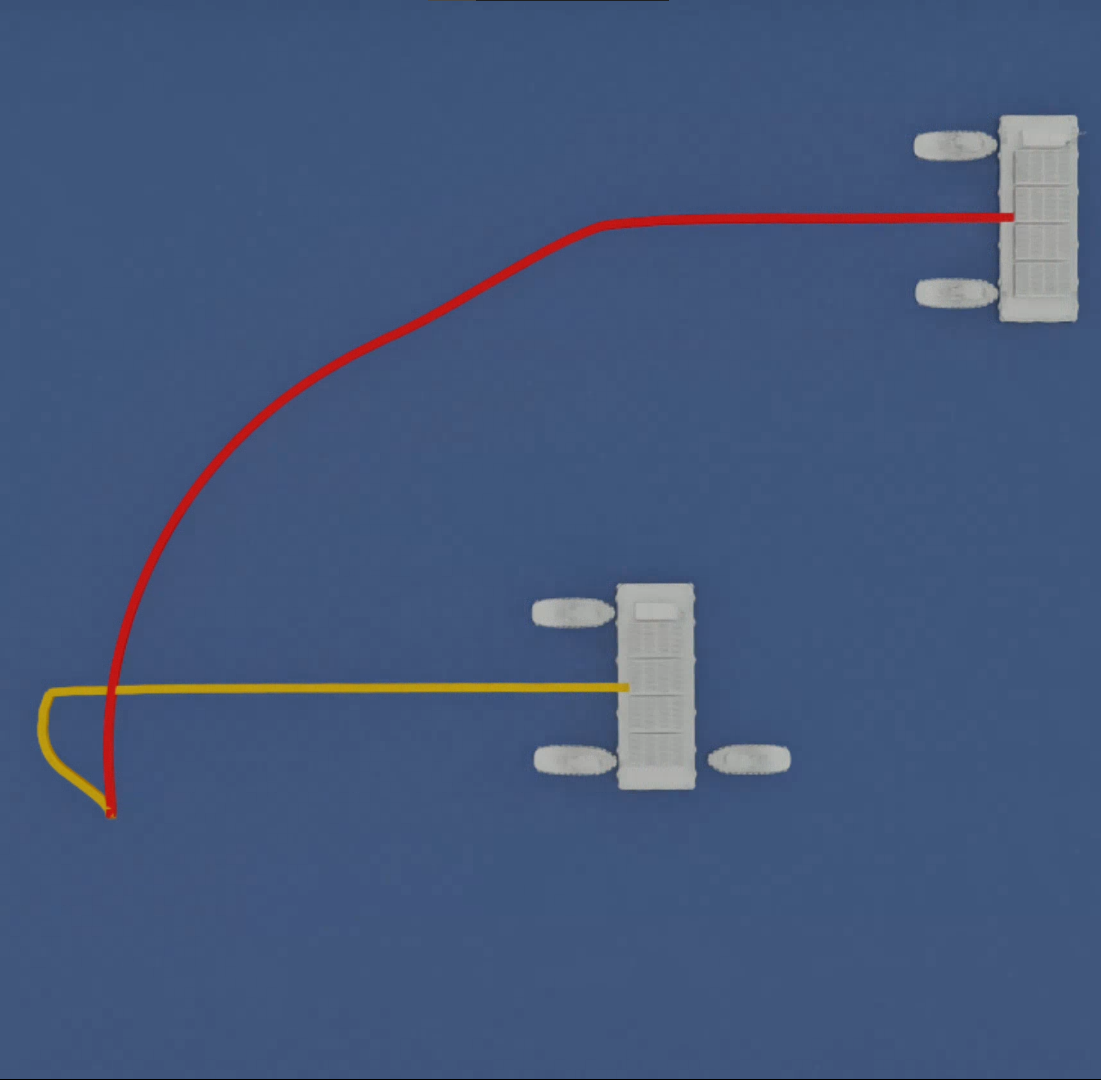}
        \caption{Three-tug turning.}
        \label{fig:three_tug}
    \end{subfigure}

    \medskip

    \begin{subfigure}{0.9\linewidth}
        \centering
        \includegraphics[width=\linewidth]{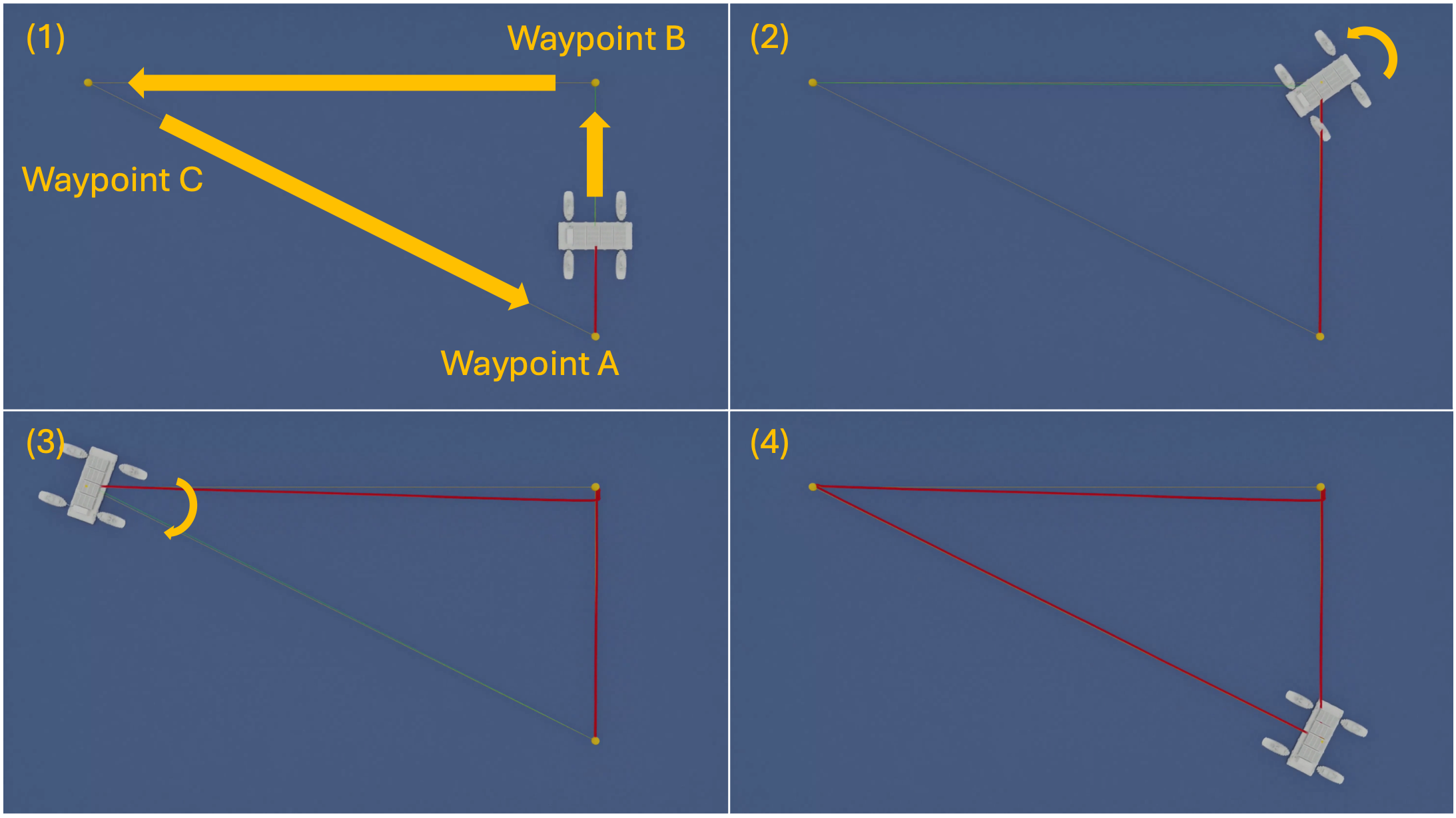}
        \caption{Four-tug full maneuver.}
        \label{fig:four_tug}
    \end{subfigure}

\caption{Zero-shot generalization to unseen tug configurations and larger team sizes.
(a) The policy trained with two tugboats is deployed to a three-tug turning configuration, reducing the lateral displacement from 111.09\,m to 21.35\,m.
(b) In the four-tug full maneuver, the barge follows waypoints B and C with in-place turning, without tugboats repositioning.}
\label{fig:generalizability}
\end{figure}

\vspace{0.2cm} \noindent \textbf{Scalability in Advanced Tasks. }
In advanced tasks, we directly deploy the policies trained with two tugboats to configurations with three and four tugboats without any additional training. As shown in Fig.~\ref{fig:generalizability}, introducing a third tugboat improves turning performance, enabling faster heading alignment and reduced deviation (reducing the lateral displacement from 111.09 m to 21.35 m). Furthermore, with four tugboats, the system executes the full maneuver sequence with higher fidelity: the barge follows the predefined waypoints more closely, performing sharper turns and reducing the shortcutting behavior observed in the two-tug configuration. This suggests that the learned behaviors generalize to larger teams and can better utilize additional actuation.

We also observe that the policy can handle unseen tug configurations, such as diagonal pushing during turning, in a zero-shot manner. In these cases, a reduced action scale is applied to ensure stable transitions between maneuver stages. This adjustment is not required for achieving the turning behavior itself, but rather for maintaining a more coherent multi-stage execution. Overall, these additional results demonstrate promising generalizability to new configurations and moderate scalability to larger teams.

\section{Conclusion}
\label{sec:conclusion}

In this work, we presented a learning-based framework for collaborative tugboat manipulation of a barge in marine environments. By combining a physics-based surface-vessel simulator with MAPPO and a structured control prior (SCP) for position-holding, the proposed method enables stable contact-based manipulation under coupled marine hydrodynamics.

Experiments on straight-line transit (SLT), turning, and deceleration show that the learned policy achieves more reliable maneuvering than the PID-based controller and centralized PPO baseline, in a more collaborative manner. The full maneuver task further demonstrates that the learned controllers can be composed across sequential command phases. The three- and four-tug experiments, meanwhile, show zero-shot generalization to unseen configurations and larger teams. These results suggest that structured multi-agent reinforcement learning is a promising direction for autonomous tugboating and collaborative marine manipulation tasks.

\clearpage
\bibliographystyle{ACM-Reference-Format}
\bibliography{bibliography}

\clearpage
\appendix

\section{Observation, Action, and Reward}
\phantomsection
\label{app:policy_spec}
This appendix summarizes the observation, action, and reward specifications used for policy training. 
The actor observation is listed in Table~\ref{tab:actor_obs}, the centralized critic state used during training is listed in Table~\ref{tab:critic_state}, and the policy action space is listed in Table~\ref{tab:action_space}. 
The reward terms used for straight-line transit and turning are summarized in Table~\ref{tab:reward_tracking_turning}, and the deceleration reward terms are summarized in Table~\ref{tab:reward_deceleration}.
\begin{table}[H]
\centering
\small
\caption{Actor observation of each tugboat agent. The same observation structure is used for straight-line transit, turning, and deceleration.}
\label{tab:actor_obs}
\renewcommand{\arraystretch}{1.08}
\begin{tabular}{
>{\raggedright\arraybackslash}p{0.30\linewidth}
>{\centering\arraybackslash}p{0.07\linewidth}
>{\raggedright\arraybackslash}p{0.57\linewidth}
}
\hline
Component & Dim. & Description \\
\hline
Previous tug-local velocity command & 3 &
Previous velocity command issued to tug \(i\),
\([v_{x,i}^{\mathrm{cmd}}, v_{y,i}^{\mathrm{cmd}}, \omega_i^{\mathrm{cmd}}]\). \\

Barge velocity & 6 &
Barge linear and angular velocity expressed in the barge reference frame. \\

Relative position error & 2 &
Planar error from tug \(i\) to its desired contact or formation target. \\

Relative heading & 2 &
Relative heading encoded as \([\sin(\psi_i-\psi_B), \cos(\psi_i-\psi_B)]\). \\

Relative angular velocity & 3 &
Angular velocity of tug \(i\) relative to the barge. \\

Relative planar velocity & 2 &
Planar velocity of tug \(i\) relative to the barge. \\

Task command & 2 &
Commanded planar barge velocity expressed in the barge reference frame. \\

Agent role & 1 &
Role indicator, \(+1\) for tug 1, \(-1\) for tug 2. \\

History & \(18H\) &
History of ego-centric proprioceptive features over \(H\) steps, the features exclude task command and agent role indicator. \\
\hline
Total & \multicolumn{2}{l}{\(18(1+H)+3 = 345\) for \(H=18\).} \\
\hline
\end{tabular}
\end{table}

\begin{table}[H]
\centering
\small
\caption{Centralized critic state used during training.}
\label{tab:critic_state}
\renewcommand{\arraystretch}{1.08}
\begin{tabular}{
>{\raggedright\arraybackslash}p{0.30\linewidth}
>{\centering\arraybackslash}p{0.07\linewidth}
>{\raggedright\arraybackslash}p{0.57\linewidth}
}
\hline
Component & Dim. & Description \\
\hline
Global state & 30 &
Previous commands, barge state, and both tugboat states arranged in self-first order. \\

Task command & 2 &
Commanded planar barge velocity. \\

Agent role & 1 &
Self-agent role indicator. \\

Privileged physics parameters & 2 &
Normalized contact-friction and barge-resistance parameters. \\

History & \(30H\) &
History of centralized features over \(H\) steps, the features exclude privileged physics parameters, task command, and agent-role indicator \\
\hline
Total & \multicolumn{2}{l}{\(30(1+H)+5 = 575\) for \(H=18\).} \\
\hline
\end{tabular}
\end{table}

\begin{table}[H]
\centering
\small
\caption{Normalized residual action space for each tugboat.}
\label{tab:action_space}
\renewcommand{\arraystretch}{1.08}
\begin{tabular}{
>{\centering\arraybackslash}p{0.24\linewidth}
>{\centering\arraybackslash}p{0.16\linewidth}
>{\raggedright\arraybackslash}p{0.42\linewidth}
}
\hline
Action & Range & Description \\
\hline
\(\Delta v_{x,i}\) & \([-1,1]\) &
Residual forward-velocity command in the tug-local frame. \\
\(\Delta \omega_i\) & \([-1,1]\) &
Residual yaw-rate command in the tug-local frame. \\
\hline
\end{tabular}
\end{table}

\begin{table}[H]
\centering
\small
\caption{Reward used for straight-line transit and turning.}
\label{tab:reward_tracking_turning}
\renewcommand{\arraystretch}{1.08}
\begin{tabular}{
>{\raggedright\arraybackslash}p{0.29\linewidth}
>{\centering\arraybackslash}p{0.13\linewidth}
>{\raggedright\arraybackslash}p{0.42\linewidth}
}
\hline
Term & Symbol & Expression \\
\hline

Velocity tracking &
\(r_v\) &
\(0.20\,\eta_v\exp(-e_v)\) \\

Barge orientation alignment &
\(r_\psi\) &
\(0.35\exp(-5.0e_\psi)\) \\

Final-goal heading alignment &
\(r_{\mathrm{goal}}\) &
\(I_{\mathrm{stg}}\,0.10\exp(-2.0e_{\mathrm{goal}})\) \\

Yaw-rate suppression &
\(r_\omega\) &
\(0.02\exp(-e_\omega)\) \\

Tug alignment &
\(r_{\mathrm{align}}\) &
\(0.15\cdot\frac{1}{2}\sum_{i=1}^{2}
\exp(-1.5e_{\mathrm{align},i})\) \\

Tug formation &
\(r_{\mathrm{form}}\) &
\(0.10\cdot\frac{1}{2}\sum_{i=1}^{2}
\exp(-2.0e_{\mathrm{form},i})\) \\

Outer-side drift suppression &
\(r_{\mathrm{drift}}\) &
\(0.18\exp(-2.0e_{\mathrm{drift}})\) \\

Outer-side velocity suppression &
\(r_{\mathrm{dir}}\) &
\(0.10\exp(-e_{\mathrm{dir}})\) \\

Early termination &
\(r_{\mathrm{term}}\) &
\(-10.0I_{\mathrm{term}}\) \\

\hline
\end{tabular}
\end{table}
Here,
\(e_v=\|\mathbf{v}_{B,xy}-\mathbf{v}_{xy}^{\mathrm{cmd}}\|/
\|\mathbf{v}_{xy}^{\mathrm{cmd}}\|\) is the relative
velocity-tracking error. The multiplier \(\eta_v\) is initialized as
\(1.0\), multiplied by \(2.0\) when the barge orientation error is below
\(3^\circ\), and further multiplied by \(1.2\) when the velocity-direction
error is below \(2^\circ\). The orientation error is
\(e_\psi=\|\mathbf{d}_B-\mathbf{d}^{\mathrm{cmd}}\|\), where
\(\mathbf{d}_B\) is the barge lateral direction and
\(\mathbf{d}^{\mathrm{cmd}}\) is the commanded velocity direction.
The final-goal heading term is used only during staged turning training:
\(I_{\mathrm{stg}}=1\) for staged turning training and \(0\) otherwise.
Its normalized error is
\(e_{\mathrm{goal}}=|\Delta\psi_{\mathrm{goal}}|/(45^\circ)\).
The yaw-rate error is
\(e_\omega=\max(|\omega_B|-\pi/180,0)\), corresponding to a threshold of
\(1^\circ/\mathrm{s}\).
For the tugboat geometry terms,
\(e_{\mathrm{align},i}\) and \(e_{\mathrm{form},i}\) denote the alignment
and formation errors of tug \(i\), with
\(e_{\mathrm{form},i}\) normalized by \(5.0\,\mathrm{m}\).
The outer-side drift error \(e_{\mathrm{drift}}\) is the outward velocity
component normalized by the commanded speed, and \(e_{\mathrm{dir}}\)
penalizes velocity directions that point outward along the barge longitudinal when the barge speed exceeds \(0.1\,\mathrm{m/s}\).
The indicator \(I_{\mathrm{term}}\in\{0,1\}\) denotes early termination.

\begin{table}[H]
\centering
\small
\caption{Reward terms used for deceleration.}
\label{tab:reward_deceleration}
\renewcommand{\arraystretch}{1.08}
\begin{tabular}{
>{\raggedright\arraybackslash}p{0.27\linewidth}
>{\centering\arraybackslash}p{0.13\linewidth}
>{\raggedright\arraybackslash}p{0.42\linewidth}
}
\hline
Term & Symbol & Expression \\
\hline

Barge speed suppression &
\(r_{\mathrm{stop},v}\) &
\(0.55\exp(-2.0e_{\mathrm{stop},v})\) \\

Brake-direction velocity tracking &
\(r_{\mathrm{brake}}\) &
\(0.06\exp(-2.0e_{\mathrm{brake}})\) \\

Relative speed suppression &
\(r_{\mathrm{rel},v}\) &
\(0.10\cdot\frac{1}{2}\sum_{i=1}^{2}
\exp(-2.5e_{\mathrm{rel},v,i})\) \\

Barge yaw-rate suppression &
\(r_{\mathrm{stop},\omega}\) &
\(0.18\exp(-1.5e_{\mathrm{stop},\omega})\) \\

Relative yaw-rate suppression &
\(r_{\mathrm{rel},\omega}\) &
\(0.07\cdot\frac{1}{2}\sum_{i=1}^{2}
\exp(-1.2e_{\mathrm{rel},\omega,i})\) \\

Tug alignment &
\(r_{\mathrm{align}}\) &
\(0.05\cdot\frac{1}{2}\sum_{i=1}^{2}
\exp(-1.5e_{\mathrm{align},i})\) \\

Tug formation &
\(r_{\mathrm{form}}\) &
\(0.07\cdot\frac{1}{2}\sum_{i=1}^{2}
\exp(-2.0e_{\mathrm{form},i})\) \\

Stop-hold bonus &
\(r_{\mathrm{hold}}\) &
\(0.20 I_{\mathrm{hold}}\) \\

Excessive deceleration penalty &
\(r_{\mathrm{decel}}\) &
\(-0.15\max((a_{\mathrm{est}}-0.40)/0.40,0)\) \\

Early termination &
\(r_{\mathrm{term}}\) &
\(-10.0 I_{\mathrm{term}}\) \\

\hline
\end{tabular}
\end{table}

Here,
\(e_{\mathrm{stop},v}=\|\mathbf{v}_{B,xy}\|/v_{\mathrm{ref}}\),
\(e_{\mathrm{brake}}=|v_{\mathrm{brake}}-v_{\mathrm{brake}}^{\mathrm{ref}}|/v_{\mathrm{ref}}\),
\(e_{\mathrm{rel},v,i}=\|\mathbf{v}_{i,xy}-\mathbf{v}_{B,xy}\|/v_{\mathrm{ref}}\),
\(e_{\mathrm{stop},\omega}=|\omega_B|/\omega_{\mathrm{ref}}\), and
\(e_{\mathrm{rel},\omega,i}=|\omega_i-\omega_B|/\omega_{\mathrm{ref}}\).
The reference speed is \(v_{\mathrm{ref}}=\max(v_0,0.5\,\mathrm{m/s})\), where \(v_0\) is the initial barge speed in the deceleration episode, and \(\omega_{\mathrm{ref}}=3^\circ/\mathrm{s}\).
The brake reference is \(v_{\mathrm{brake}}^{\mathrm{ref}}=\max(v_0-0.33t,0)\).
The estimated deceleration \(a_{\mathrm{est}}\) is computed using an exponential moving average with smoothing factor \(0.25\).
The stop-hold indicator \(I_{\mathrm{hold}}\) becomes one after the stopping condition is satisfied for 20 consecutive control steps.
The indicator \(I_{\mathrm{term}}\in\{0,1\}\) denotes early termination.

\section{Simulator Details}
\subsection{Details of Wave Model}
\phantomsection
\label{app:wave}
The simplified height-field model used in Section~\ref{sec:simulation-framework} is based on the classical Gerstner-wave formulation~\cite{von1804theorie,fournier1986simple}. In the classical formulation, the water surface is described in Lagrangian form as
\begin{equation}
\mathbf{p}' = \mathbf{p} + \sum_k a_k \mathbf{d}_k \cos(\mathbf{k}_k \cdot \mathbf{p} - \omega_k t), \quad
z = \sum_k a_k \sin(\mathbf{k}_k \cdot \mathbf{p} - \omega_k t),
\end{equation}
where $\mathbf{p} = (x,y)^T$ is the horizontal position, $a_k$, $\mathbf{d}_k$, $\mathbf{k}_k$, and $\omega_k$ denote the amplitude, propagation direction, wave vector, and angular frequency of each component.

In this formulation, the horizontal position $\mathbf{p}'$ depends on the wave displacement, meaning that the surface is implicitly defined. As a result, querying the water height at a fixed spatial location requires solving a nonlinear (transcendental) equation.

To enable efficient wave-height queries, we simplify the model by neglecting horizontal particle displacement and retaining only the vertical component, yielding an explicit surface representation:
\begin{equation}
\eta(\mathbf{p}, t)
=
z_w
+
\frac{A}{N_d}
\sum_{j=1}^{N_d}
\alpha_j
\sum_{m=1}^{N_\omega}
a_m
\cos\!\left(
k_m \, \mathbf{d}_j^T \mathbf{p}
-
\omega_m t
+
\phi_m
\right),
\end{equation}
where $\mathbf{p} = (x,y)^T$ is the horizontal position, $z_w$ denotes the mean water level, $A$ is a global amplitude scaling factor, $N_d$ and $N_\omega$ are the number of sampled propagation directions and frequency components respectively, $\mathbf{d}_j$ are unit direction vectors, $\alpha_j$ controls directional spreading, $a_m$ are spectral amplitudes, $k_m$ and $\omega_m$ denote the wave number and angular frequency, and $\phi_m$ is a phase offset. The spectral amplitudes are derived from a TMA spectrum~\cite{bouws1985similarity}, a finite-depth extension of the JONSWAP model~\cite{hasselmann1973measurements}, allowing the wave field to reflect wind-driven sea states while accounting for water depth effects.

\begin{figure}
    \centering
    \includegraphics[width=1\linewidth]{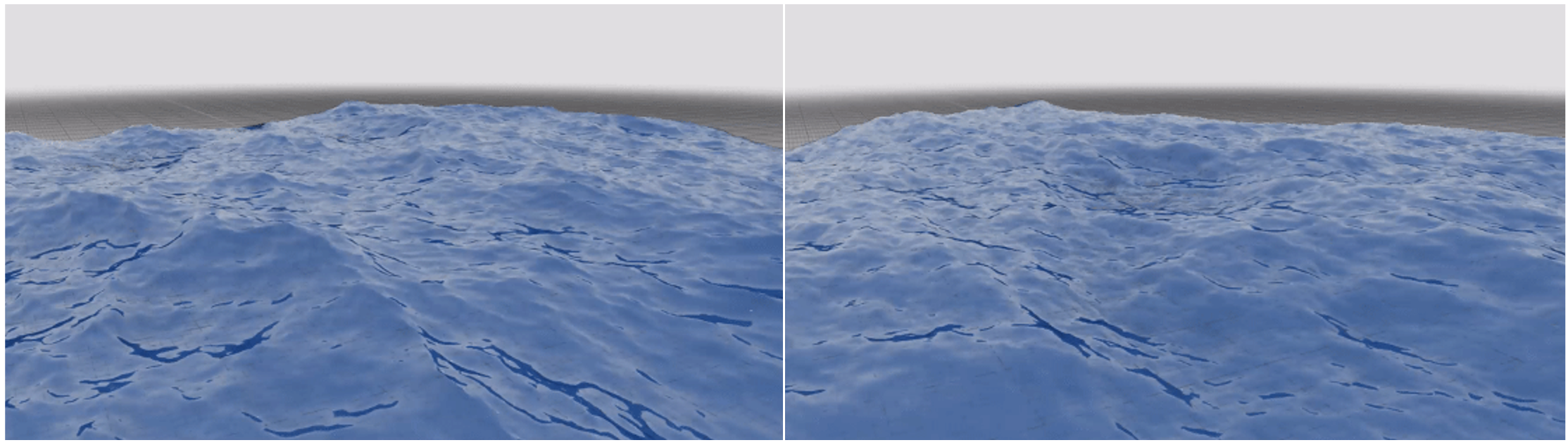}
    \caption{Nominal Gerstner-wave (left) and simplified Gerstner-wave model (right, ours).}
    \label{fig:wave}
\end{figure}

This formulation allows direct evaluation of surface height at arbitrary spatial positions without solving implicit equations. The removal of horizontal displacement introduces only minor discrepancies in surface geometry while significantly simplifying computation. Prior work suggests that horizontal particle motion mainly serves to enhance visual features such as crest sharpening~\cite{fernando2004gpu}, while certain established wave models adopt height-field representations without explicit horizontal displacement~\cite{tessendorf2001simulating}. This supports the validity of removing the horizontal component in our formulation. As illustrated in Fig.~\ref{fig:wave}, the resulting surface can capture the majority of the water height variation, only losing trivial details like the sharpness of the crest. 

\subsection{Details of the Buoyancy Calculation}
\phantomsection
\label{app:buoyancy}
The buoyancy calculation is based on a voxelized approximation of the vessel geometry. Each vessel is discretized into a set of volumetric cuboids, and buoyancy is computed by estimating the displaced water volume from the submerged voxels.

For each voxel $i$, a binary submerged indicator $s_i \in \{0,1\}$ is determined by comparing the voxel-center height with the local water surface height:
\begin{equation}
s_i =
\left\{
\begin{array}{ll}
1 & \quad z_i < \eta(x_i, y_i, t) \\
0 & \quad \mathrm{otherwise},
\end{array}
\right.
\end{equation}
where $(x_i,y_i,z_i)$ denotes the voxel-center position in the world frame, and $\eta(x,y,t)$ is the wave surface elevation at horizontal position $(x,y)$ and time $t$.

The buoyancy contributed by voxel $i$ is computed according to Archimedes' principle:
\begin{equation}
\mathbf{f}_i = \rho_w g V_i s_i \mathbf{n}_z,
\end{equation}
where $\rho_w$ is the water density, $g$ is the gravitational acceleration, $V_i$ is the voxel volume, and $\mathbf{n}_z$ is the upward unit vector.

The total buoyancy is obtained by summing the contributions of all voxels:
\begin{equation}
\mathbf{F}_b = \sum_i \mathbf{f}_i.
\end{equation}

The center of buoyancy is defined as the volume-weighted centroid of the submerged voxels:
\begin{equation}
\mathbf{r}_c =
\frac{\sum_i s_i V_i \mathbf{r}_i}
{\sum_i s_i V_i},
\end{equation}
where $\mathbf{r}_i$ is the position of voxel $i$.

The buoyancy is applied at $\mathbf{r}_c$. The corresponding torque about the vessel center of mass $\mathbf{r}_{\mathrm{com}}$ is given by
\begin{equation}
\boldsymbol{\tau}_b =
(\mathbf{r}_c - \mathbf{r}_{\mathrm{com}}) \times \mathbf{F}_b .
\end{equation}

\begin{figure}
    \centering
    \includegraphics[width=0.9\linewidth]{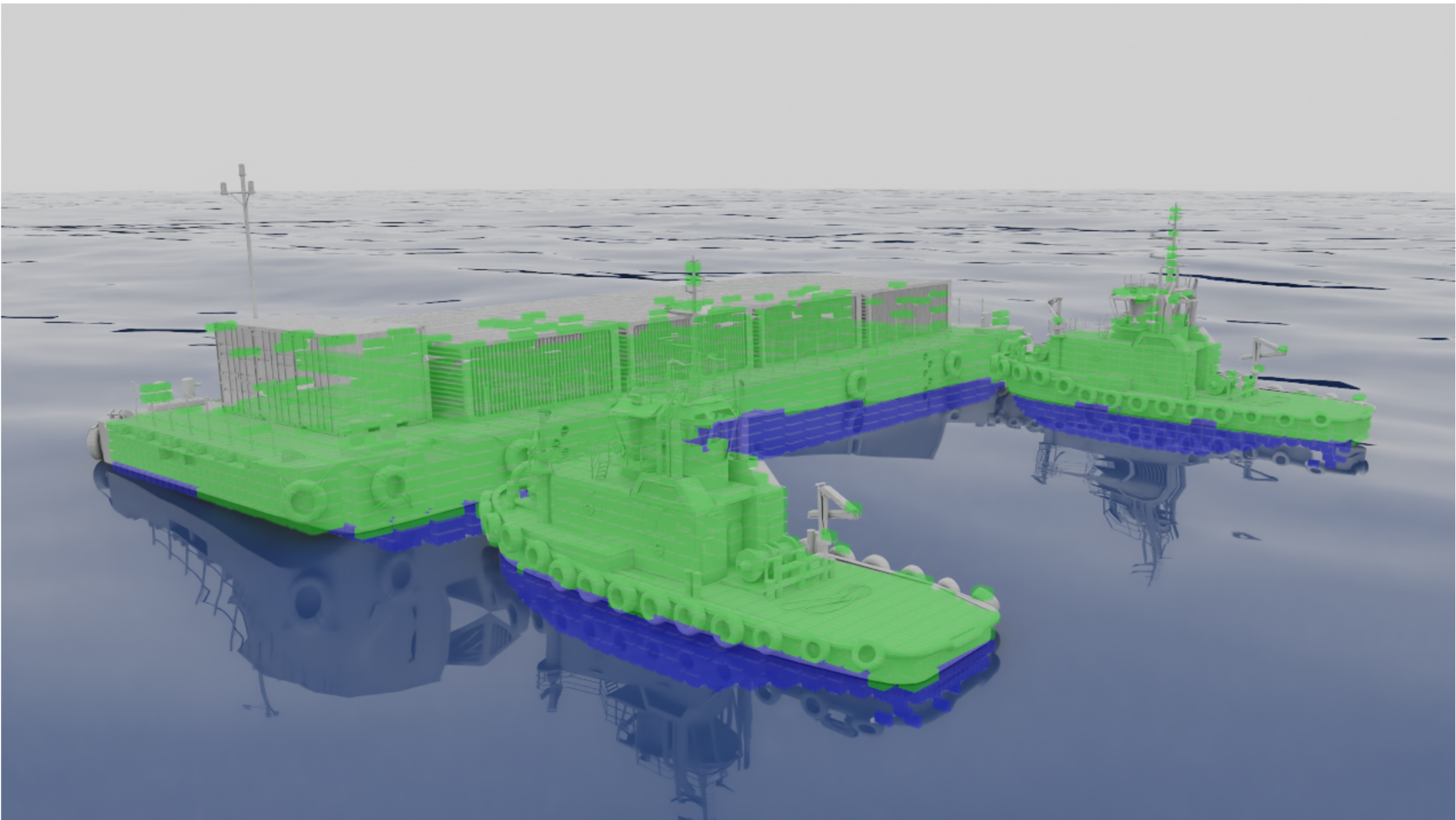}
    \caption{Voxelization of barge and tugboats in our simulator: the green voxels represent unsubmerged parts, and blue voxels represent the submerged parts.}
    \label{fig:voxelization}
\end{figure}

This voxel-based formulation is fully parallelizable and integrates efficiently with GPU-based simulation. The voxel resolution provides a controllable trade-off between computational cost and buoyancy accuracy. The voxelization of barge and tugboats in the simulator can be seen in Fig~\ref{fig:voxelization}.

\subsection{Details of the Hydrodynamic Resistance Model}
\phantomsection
\label{app:Hydrodynamics Resistance Model}
For the barge, the hull hydrodynamics follow the unified hull-force formulation~\cite{yoshimura2009unified}, in which the surge, sway, and yaw components are represented by a combination of linear hydrodynamic derivatives and cross-flow drag terms. 
Let $u$, $v$, and $r$ denote the surge velocity, sway velocity, and yaw rate in the body-fixed frame, respectively. The planar speed magnitude is defined as
\begin{equation}
U = \sqrt{u^2 + v^2},
\end{equation}
and the corresponding non-dimensional velocities are
\begin{equation}
u' = \frac{u}{U}, \qquad
v' = \frac{v}{U}, \qquad
r' = \frac{r L_{pp}}{U},
\end{equation}
where $L_{pp}$ is the length between perpendiculars. The hull surge force, sway force, and yaw moment are then written as
\begin{equation}
X_H = X_{\mathrm{cor}} \, \frac{1}{2} \rho L_{pp} d U^2 X_H',
\end{equation}
\begin{equation}
Y_H = Y_{\mathrm{cor}} \, \frac{1}{2} \rho L_{pp} d U^2 Y_H',
\end{equation}
\begin{equation}
N_H = N_{\mathrm{cor}} \, \frac{1}{2} \rho L_{pp}^2 d U^2 N_H',
\end{equation}
where $\rho$ is the water density and $d$ is the draft. The corresponding non-dimensional hull-force expressions are
\begin{equation}
X_H' = - X_0' u' + \left( m_y' + X_{vr}' \right) v' r',
\end{equation}
\begin{equation}
Y_H' = Y_v' v' \left| u' \right| + \left( Y_r' - m_x' \right) r' u'
- C_D \int_{-0.5}^{0.5} \left( v' + C_{rY} r' x \right)
\left| v' + C_{rY} r' x \right| \, \mathrm{d}x,
\end{equation}
\begin{equation}
N_H' = N_v' v' u' + N_r' r' \left| u' \right|
- C_D \int_{-0.5}^{0.5} x \left( v' + C_{rN} r' x \right)
\left| v' + C_{rN} r' x \right| \, \mathrm{d}x.
\end{equation}
This structure preserves the conventional linear maneuvering derivatives while augmenting the sway and yaw components with cross-flow drag contributions. The primed quantities $m_x'$ and $m_y'$ denote the non-dimensional added-mass contributions, whereas the unprimed quantities $m_x$, $m_y$, and $J_{zz}$ introduced below denote the dimensional added masses and added yaw inertia used in the simulator implementation.

In the original unified/MMG-style equations, added masses and added yaw inertia enter the inertia terms~\cite{yasukawa2015introduction}. In the present simulator implementation, these effects are incorporated through equivalent correction factors applied to the hydrodynamic force and moment terms, thereby retaining the effective increase in inertia without explicitly reformulating the mass matrix. Specifically,
\begin{equation}
X_{\mathrm{cor}} = \frac{m}{m + m_x}, \qquad
Y_{\mathrm{cor}} = \frac{m}{m + m_y}, \qquad
N_{\mathrm{cor}} = \frac{I_{zz}}{I_{zz} + J_{zz}},
\end{equation}
where $m$ is the barge mass, $I_{zz}$ is the yaw moment of inertia, and $m_x$, $m_y$, and $J_{zz}$ are estimated using empirical regressions based on principal dimensions and block coefficient~\cite{zhang2023twin,zhou1983manoeuvring}.

For resistance coefficient $X_0'$, the unified model recommends prediction by conventional powering methods. Accordingly, we estimate $X_0'$ through a conventional resistance-decomposition approach~\cite{birk2019fundamentals,ittc_1978_ppm} and convert it into the normalization used by the unified hull-force model:
\begin{equation}
X_0' = \left[ \left( 1 + K \right) C_F + \Delta C_{F} + C_{AAS} \right]
\frac{S_{\mathrm{wet}}}{L_{pp} d},
\end{equation}
where $S_{\mathrm{wet}}$ is the wetted surface area and $K$ is the form factor. The friction coefficient $C_F$ is computed using the ITTC-1957 friction line~\cite{birk2019fundamentals}, given by
\begin{equation}
C_F = \frac{0.075}{\left( \log_{10} Re - 2 \right)^2},
\end{equation}
and the roughness allowance $\Delta C_{F}$ is adopted from the ITTC-1978 performance prediction method~\cite{ittc_1978_ppm},
\begin{equation}
\Delta C_{F} = \left[ 105 \left( \frac{k_s}{L_{WL}} \right)^{1/3} - 0.64 \right] \times 10^{-3},
\end{equation}
where $Re$ is the Reynolds number, $k_s$ is the equivalent sand roughness, and $L_{WL} \approx L_{pp}$ is used in the present implementation. The form factor $K$ is approximated using stern-shape parameters following a Kijima-type approximation~\cite{maimun2011manoeuvring}, and $C_{AAS}$ is introduced as a small empirical air-resistance allowance~\cite{birk2019fundamentals}. 

Since the simulator focuses on low-speed barge maneuvering, with a typical barge speed of about \(1\,\mathrm{m/s}\), corresponding to a low-Froude-number regime for the present hull size, the resistance is assumed to be dominated by viscous-related contributions, while wave-making effects are treated as secondary, so wave resistance is not modeled explicitly. Likewise, appendage resistance is not parameterized separately because the present model is formulated as a baseline barge-hull resistance model.

\begin{figure}[t]
    \centering
    \begin{subfigure}{\linewidth}
        \centering
        \includegraphics[width=0.9\linewidth]{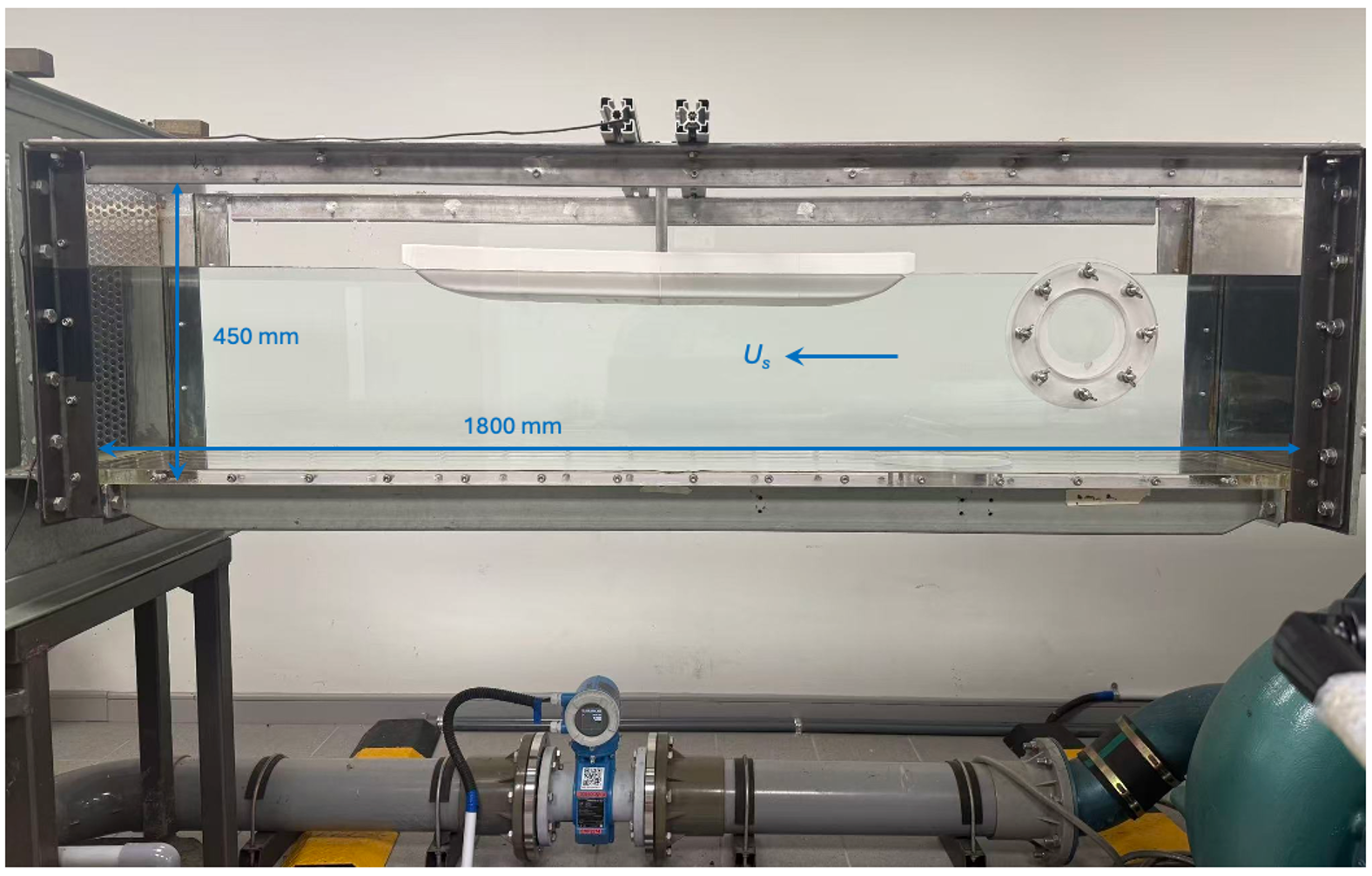}
        \caption{Water-tunnel setup.}
        \label{fig:water_tunnel_setup}
    \end{subfigure}

    \vspace{1ex}

    \begin{subfigure}{\linewidth}
        \centering
        \includegraphics[width=\linewidth]{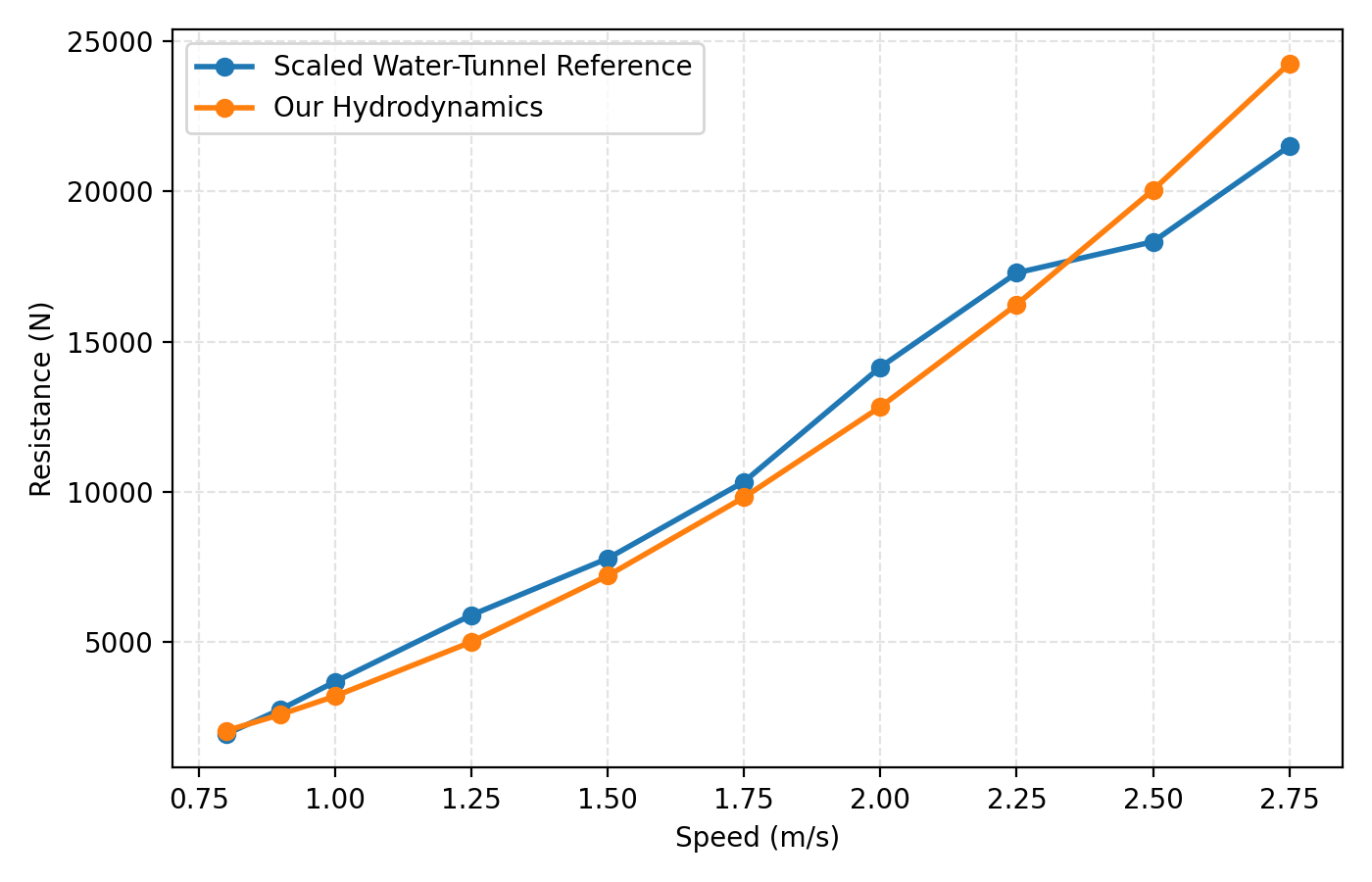}
        \caption{Resistance comparison.}
        \label{fig:resistance_comparison}
    \end{subfigure}
    \caption{Water-tunnel calibration of the barge resistance model. 
    (a) Experimental setup with a geometrically scaled barge model. 
    (b) Comparison between the scaled water-tunnel reference and the simulator hydrodynamic resistance after applying the empirical correction factor of 2.5.}
    \label{fig:water_tunnel_calibration}
\end{figure}

This hydrodynamic resistance model combines unified-model terms with several empirical coefficients and should therefore be regarded as a practical low-speed modeling component rather than a fully predictive hydrodynamic solver. 
To provide an experimental reference for the resistance magnitude, we conducted a small-scale water-tunnel test using a geometrically scaled barge model, as shown in Fig.~\ref{fig:water_tunnel_setup}. 
The measured model-scale resistance was extrapolated to an approximate full-scale reference using a Froude-similarity-inspired scaling procedure. 
This extrapolated result is used as a trend-level calibration reference for the simulator resistance, rather than as a strict full-scale hydrodynamic validation.

As shown in Fig.~\ref{fig:resistance_comparison}, after applying the global correction factor of 2.5, the simulator resistance shows close agreement with the scaled water-tunnel reference over the low-speed maneuvering regime. 
We therefore apply this empirical correction to improve the RL-oriented fidelity of the physics-based simulator.

\section{Detailed Experimental Settings}
\phantomsection
\label{sec:exp_setting}
This appendix provides task parameters used in the experiments.

\begin{table*}[t]
\centering
\caption{Task settings for the basic tasks.}
\label{tab:task_settings}
\small
\setlength{\tabcolsep}{4pt}
\renewcommand{\arraystretch}{1.2}
\renewcommand{\tabularxcolumn}[1]{m{#1}}

\begin{tabularx}{\textwidth}{
>{\centering\arraybackslash}m{0.155\textwidth}|
>{\centering\arraybackslash}m{0.015\textwidth}|
>{\centering\arraybackslash}m{0.11\textwidth}
>{\centering\arraybackslash}m{0.11\textwidth}
>{\centering\arraybackslash}m{0.11\textwidth}
>{\centering\arraybackslash}X
>{\centering\arraybackslash}m{0.09\textwidth}
>{\centering\arraybackslash}m{0.11\textwidth}}
\hline
\multicolumn{2}{c|}{Task} & Command Angle ($^\circ$) & Initial Velocity (m/s) & Init. Vel. Direction ($^\circ$) & Overshoot Factor & Wave (m) & Task Horizon (s) \\
\hline

\multirow{2}{*}{\parbox{0.12\textwidth}{\centering Straight}}
& \hypertarget{tab-task-a}{\hyperlink{text-task-a}{A}} & 0 & 1 & 0 & N/A & 0 & 60 \\
\cline{2-8}
& \hypertarget{tab-task-b}{\hyperlink{text-task-b}{B}} & 0 & 1 & N/A & N/A & 0 & 60 \\
\hline

\multirow{2}{*}{\parbox{0.12\textwidth}{\centering Turning}}
& \hypertarget{tab-task-c}{\hyperlink{text-task-c}{C}} & -90 & 1 & 0 & PID 40\%, RL 25\% & 0 & 150 \\
\cline{2-8}
& \hypertarget{tab-task-d}{\hyperlink{text-task-d}{D}} & -30 & 1 & 0 & 0\% & 0 & 300 \\
\hline

\multirow{2}{*}{\parbox{0.12\textwidth}{\centering Deceleration}}
& \hypertarget{tab-task-e}{\hyperlink{text-task-e}{E}} & N/A & 1 & 180 & N/A & 0 & 2 \\
\cline{2-8}
& \hypertarget{tab-task-f}{\hyperlink{text-task-f}{F}} & N/A & 1 & 195 & N/A & 0 & 2 \\
\hline
\end{tabularx}
\end{table*}

\begin{table}[H]
\centering
\caption{Proportional gains used by the PID-based controller for each basic task.}
\label{tab:pid_gains}
\begin{tabularx}{\linewidth}{l *{3}{>{\centering\arraybackslash}X}}
\hline
Task & $k_{\mathrm{yaw}}$ & $k_{\mathrm{velocity-direction}}$ & $k_{\mathrm{velocity-magnitude}}$ \\
\hline
SLT & 5.0 & -5.0 & 10.0 \\
Turning & 2.0 & -0.25 & 0.25 \\
Deceleration & 0.0 & 0.0 & 1.0 \\
\hline
\end{tabularx}
\end{table}

\begin{table}[H]
\caption{Gain settings for the \(x\)- and yaw-axis SCP strength ablation. Derivative gains are fixed to zero for all variants.}
\label{tab:controller_ablation}
\centering
\footnotesize
\setlength{\tabcolsep}{4pt}
\renewcommand{\arraystretch}{1.1}
\begin{tabularx}{\linewidth}{>{\raggedright\arraybackslash}X|ccc}
\hline
\multicolumn{1}{c|}{Variant} & $x$ $K_p$ & $y$ $K_p$ & yaw $K_p$ \\
\hline
MAPPO Policy + Full SCP             & 5.0 & 5.0 & 5.0 \\
MAPPO Policy + Weak SCP        & 2.5 & 5.0 & 2.5 \\
MAPPO Policy without SCP       & 0.0 & 5.0 & 0.0 \\
MAPPO Policy with decaying SCP & $5.0\rightarrow0.0^{*}$ & 5.0 & $5.0\rightarrow0.0^{*}$ \\
\hline
\end{tabularx}

\vspace{1mm}
\begin{minipage}{0.86\linewidth}
\footnotesize
\(^*\) The \(x\)- and yaw-axis proportional gains are decayed during training and set to 0.0 at inference. 
All derivative gains are fixed to 0.0.
\end{minipage}
\end{table}

\end{document}